\pdfoutput=1

\documentclass[11pt]{article}

\usepackage{multirow} 
\usepackage[most]{tcolorbox} 
\usepackage{xcolor} 
\usepackage[final]{acl}

\usepackage{times}
\usepackage{latexsym}
\usepackage{booktabs}
\usepackage{xcolor}
\usepackage{lipsum}  
\usepackage{amssymb}
\usepackage{url}

\usepackage[T1]{fontenc}

\usepackage[utf8]{inputenc}

\usepackage{microtype}

\usepackage{inconsolata}

\usepackage{graphicx}

\usepackage{color} 

\usepackage[most]{tcolorbox}
\usepackage{listings}
\usepackage{float}

\tcbset{
  aibox/.style={
    width=220pt,
    top=10pt,
    colback=white,
    colframe=black,
    colbacktitle=black,
    enhanced,
    center,
    attach boxed title to top left={yshift=-0.1in,xshift=0.15in},
    boxed title style={boxrule=0pt,colframe=white,},
  }
}
\newtcolorbox{AIbox}[2][]{aibox,title=#2,#1}

\title{How Well Do Multi-modal LLMs Interpret CT Scans? \\ An Auto-Evaluation Framework for Analyses}

\author{
 \textbf{Qingqing Zhu\textsuperscript{1,*}},
 \textbf{Benjamin Hou\textsuperscript{2,5,*}},
 \textbf{Tejas Sudarshan Mathai\textsuperscript{2}},
 \textbf{Pritam Mukherjee\textsuperscript{2}},
\\
 \textbf{Qiao Jin\textsuperscript{1}}, 
 \textbf{Xiuying Chen\textsuperscript{4}},
 \textbf{Zhizheng Wang\textsuperscript{1}},
 \textbf{Ruida Cheng\textsuperscript{3}},
\\
 \textbf{Ronald M. Summers\textsuperscript{2,**}},
 \textbf{Zhiyong Lu\textsuperscript{1,**}}
\\
 \textsuperscript{1}National Center for Biotechnology Information, National Library of Medicine, \\
 \textsuperscript{2}Imaging Biomarkers and Computer-Aided Diagnosis Laboratory, Clinical Center, \\
 \textsuperscript{3}Center for Information Technology, National Institutes of Health, \\
 \textsuperscript{4}King Abdullah University of Science \& Technology, \\
 \textsuperscript{5}Biomedical Image Analysis, Imperial College London 
\\
$^{*}$Joint co-authors, $^{**}$Joint senior co-authors
}

\begin{document}
\maketitle
\begin{abstract}
Automatically interpreting CT scans can ease the workload of radiologists. However, this is challenging mainly due to the scarcity of adequate datasets and reference standards for evaluation.  This study aims to bridge this gap by introducing a novel evaluation framework, named ``GPTRadScore''. This framework assesses the capabilities of multi-modal LLMs, such as GPT-4 with Vision (GPT-4V), Gemini Pro Vision, LLaVA-Med, and RadFM, in generating descriptions for prospectively-identified findings. 
By employing a decomposition technique based on GPT-4, GPTRadScore compares these generated descriptions with gold-standard report sentences, analyzing their accuracy in terms of body part, location, and type of finding.
 Evaluations demonstrated a high correlation with clinician assessments and highlighted its potential over traditional metrics, such as BLEU, METEOR, and ROUGE. 
 Furthermore, to contribute to future studies, we plan to release a benchmark dataset annotated by clinicians.
 Using GPTRadScore, we found that while GPT-4V and Gemini Pro Vision fare better, their performance revealed significant areas for improvement, primarily due to limitations in the dataset used for training these models. To demonstrate this potential, RadFM was fine-tuned and it resulted in significant accuracy improvements: location accuracy rose from 3.41\% to 12.8\%, body part accuracy from 29.12\% to 53\%, and type accuracy from 9.24\% to 30\%, thereby validating our hypothesis.

\end{abstract}


\section{Introduction}
\label{sec_Intro}

In current clinical practice, a radiologist communicates the results of an imaging exam for a patient to their referring doctor through a signed report. While reading the patient exam, the radiologist routinely use Speech Recognition Software (SRS) that converts dictated speech into text. SRS has significantly reduced the report turn-around time. However, any errors resulting from the dictation have to be corrected by the radiologists themselves, and persistent errors can negatively impact the interpretation of patient diagnoses and can have medico-legal ramifications \citep{Smith2001_legalRads}. These errors are most common for cross-sectional imaging \cite{Ringler2017_errorsDictation}, such as CT and MR, and the volume of these exams has steadily increased each year \cite{Mahesh2023}. This has led to a 54-72\% radiologist burn-out rate \cite{Fawzy2023} where they are under increased pressure to deal with a substantially higher number of patients while maintaining a high level of accuracy.

To ameliorate the radiologist workload, various transformer-based approaches have been proposed to generate radiology reports in one shot \cite{chen-etal-2020-generating,chen-etal-2021-cross-modal}. However, these efforts focus mainly on chest radiographs (CXR), with limited attention to CT \cite{Ichinose2023}. Developing CT-based reporting methods is challenging due to the 3D nature of CT data, computational complexity, and the factual accuracy of reporting needed. Recent advances with Large Language Models (LLMs) like GPT-4 \cite{achiam2023gpt}, GPT-4 Vision (GPT-4V), Gemini Pro Vision \cite{team2023gemini}, LLaVA-Med \cite{li2024llava}, and Radiology Foundation Model (RadFM) \cite{wu2023generalist} show potential for various tasks, such as taking medical exams, note-taking, and disease diagnosis \cite{tian2024opportunities,nori2023capabilities,jin2023retrieve}. These multi-modal models could pre-fill the ``findings'' section of radiology reports for quick review by radiologists \cite{ZhuMMPSL23}.




Despite these advances, crucial factors determining their clinical use involve: (1) radiologist trust, and (2) easy interpretation and evaluation of the generated content. Current evaluation metrics, including Natural Language Generation (NLG) and Clinical Efficacy (CE) metrics, are notoriously limited \cite{irvin2019chexpert,ZhuMMPSL23,zhu2024leveraging,jin2024hidden} when it comes to capturing the semantic richness and clinical relevance necessary for radiology reports. Additionally, they lack the explanatory power that is required for clinical use. 

In this paper, we present a novel evaluation framework to assess the capability of multi-modal LLMs to generate diagnostically accurate descriptions of CT-based findings for radiology reports. CT slices with an abnormal finding were fed to a multi-modal LLM (e.g., GPT-4V) that generated a description of the abnormality. A language-centric GPT-4 model decomposed the summary into its characteristics (body part, location, type), evaluated them against gold-standard references, and scored the description based on its clinical relevance and accuracy. Our contribution can be summarized as follows:

    \noindent $\bullet$ 
    We introduced a new framework named ``GPTRadScore'', designed to evaluate the accuracy of multi-modal LLMs in describing CT scan findings, specifically focusing on the precision of identifying body parts, locations, and types of findings.
    
    \noindent $\bullet$  To validate this approach, we conducted human evaluations on 500 cases in collaboration with clinicians. Furthermore, we intend to publicly release these expert annotations (with CC-BY-NC-SA 4.0 licence) to establish a new benchmark for accuracy in future assessments.
    
   \noindent $\bullet$  Four recent multi-modal LLMs were evaluated for their ability to describe CT findings. 
    
    \noindent $\bullet$  RadFM was fine-tuned with domain-specific data to improve its generation accuracy.

\section{Related Works}

Early efforts in extracting pathologies utilized NLG rules, which were crafted to isolate specific disease features. Notable examples include the cheXpert-labeler and NegBio \cite{wang2017chestx, peng2018negbio, irvin2019chexpert}, both of which were employed to derive disease labels in chest X-rays. With the advent of transformer models, notably the BERT model, a more advanced solution, the cheXbert-labeler, was introduced. The cheXbert-labeler is a model specifically trained on the CheXpert dataset to perform this task.

As LLMs gain popularity, their integration into radiology becomes increasingly inevitable. These models, including multimodal LLMs, are set to assist in clinical decisions, extract information from clinical notes, and generate radiological reports, showcasing their broad utility in the field \cite{zhou2023skingpt,zhou2023path,bhayana2024chatbots,tian2024opportunities}. 

LLMs possess the advanced capability for complex reasoning, making them highly suitable for analyzing AI-generated radiological reports in comparison to ground truth. Leveraging LLMs to evaluate radiological reports harnesses their analytical power and provides a scalable solution for managing large datasets, potentially containing thousands of reports. Relying on clinicians to validate these reports is an inefficient use of their time, given their essential roles in direct patient care and decision-making. By using LLMs for initial evaluations, healthcare systems can reserve clinicians' expertise for tasks where human judgment is crucial, optimizing resources and potentially speeding up the diagnostic process.


\citet{wang2024llm} recently introduced LLM-RadJudge, a method that compares the performance of various LLMs and demonstrates that using GPT-4, their proposed metric achieves evaluation consistency close to that of radiologists. Furthermore, they constructed a dataset based on LLM evaluation results and used knowledge distillation to train a smaller model, which achieves evaluation capabilities comparable to GPT-4. Similarly, \citet{liu2024mrscore} proposed MRScore, a framework akin to LLM-RadJudge. \citet{zhu2024leveraging} proposed a method that combines the expertise of professional radiologists with LLMs such as GPT-3.5 and GPT-4. Using In-Context Instruction Learning (ICIL) and Chain of Thought (CoT) reasoning to align LLM evaluations with radiologist standards, experimental results demonstrated greater alignment with expert evaluations, surpassing traditional NLG metrics such as BLEU, ROUGE, and METEOR.

Despite these advancements, there is still no automated system for validating the clinical accuracy of CT reports, largely because of the scarcity of high-quality datasets and the complexity of CT imaging, which involves a broad range of body parts and requires extensive anatomical knowledge. This paper introduces GPTRadScore, a novel evaluation framework that assesses the capabilities of multimodal LLMs. It uses a decomposition method based on GPT-4, which mimics clinicians' evaluation processes, comparing AI-generated descriptions with the actual ground truth across factors such as body part, location, and type.

\begin{figure*}[!t]
  \centering
  \includegraphics[width=0.95\textwidth]{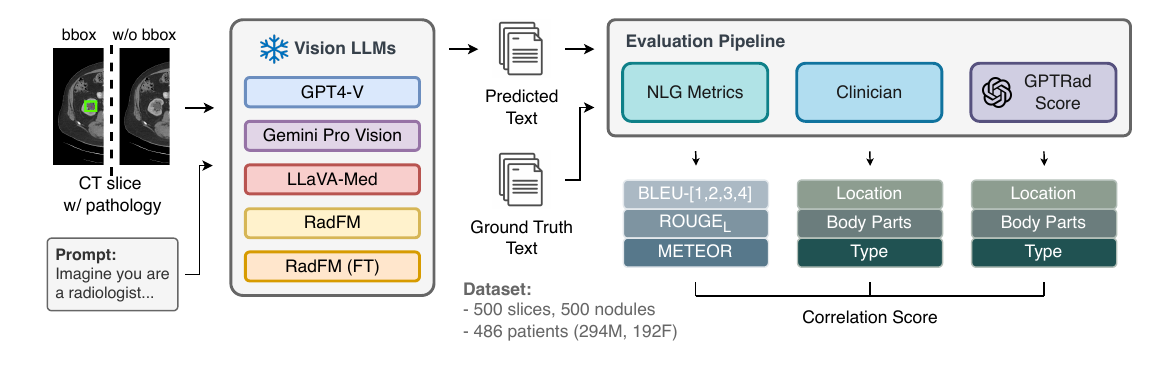}
  \caption{``GPTRadScore'' framework for the auto-evaluation of LLM descriptions of CT-based findings. CT slices with outlined lesions were fed to vision-based LLMs that generated a description of the finding. They were then evaluated against the gold-standard sentences by a clinician, with NLG metrics, and auto-evaluation with GPT-4.}
  \label{fig:workflow}
\end{figure*}

\section{Methods}

This study introduces a novel ``GPTRadScore'' framework for evaluating the accuracy of multi-modal LLMs in generating clinical descriptions of CT-based findings. Figure \ref{fig:workflow} illustrates the experimental design. We break down the experimental setting into two integral steps:
1. Generating Descriptions of CT Findings: (1)Visual Context Integration: CT slices with abnormalities are marked with bounding boxes to provide clear visual context to the multi-modal LLMs.
(2) Text-Based Chain-of-Thought (CoT): The multi-modal LLMs generate free-text descriptions of the abnormalities, focusing on body part, specific location, and type of finding.
(3) Fine-Tuning RadFM: RadFM was fine-tuned using domain-specific data from the DeepLesion dataset to improve its accuracy in generating clinically relevant descriptions of findings.
2. Evaluation Process (GPTRadScore): GPT-4 was used to compare the generated descriptions against gold-standard report sentences. 
Scores were assigned based on clinical relevance and accuracy, mimicking clinician assessment.

\label{sec_Methods}

\noindent
\subsection{Dataset}
To the best of our knowledge, no publicly available dataset pairs CT exams with corresponding radiology reports for lesions. For this retrospective study, the DeepLesion dataset \cite{yan2018deeplesion} was utilized. The dataset comprises 23,436 CT slices and 8,340 studies with reports from 3,832 patients (mean age: 51, s.d.: 17; 2,085 males). Report sentences containing prospective RECIST-based measurements, made by radiologists and referred to as ``bookmarks'', were extracted using regular expressions. An enclosed bounding box was also created from the prospective measurement to highlight the finding in the CT slice.

The main portion of this experiment utilizes a subset of the DeepLesion dataset \cite{yan2019holistic}, which comprises 496 CT volumes (496 studies) from 486 patients (mean age: 52.2, s.d.: 17.7; 294 males). The subset contained 500 lesions of various kinds (e.g., liver, kidney, bone, etc.) that were prospectively marked in 500 CT slices. The subset also provided specific characteristics of lesions that were extracted from the sentences in the radiology reports using an automated method. These included the body part where the lesion is located, the fine-grain location within that region (e.g., upper pole of left kidney), and the type of lesion attributes. As certain lesion characteristics were missed by the automated extraction, two board-certified radiologists, each with 10+ years of experience, manually reviewed and comprehensively annotated any missing lesion characteristics.

\subsection{Visual Context Integration}

In addition to the CT slice that shows the abnormal finding, a visual prompt was also provided. This prompt, in the form of a bounding box, delineated the abnormality before being input into the multi-modal LLM. 
The clear visual context was hypothesized to enhance the accuracy of the generated descriptions of findings.

\begin{figure}[!ht]
\small
\begin{tcolorbox}[colback=white, sharp corners, boxrule=0.5pt, colframe=black, title=Case Study]
\begin{center} 
\includegraphics[width=4cm,height=4cm]{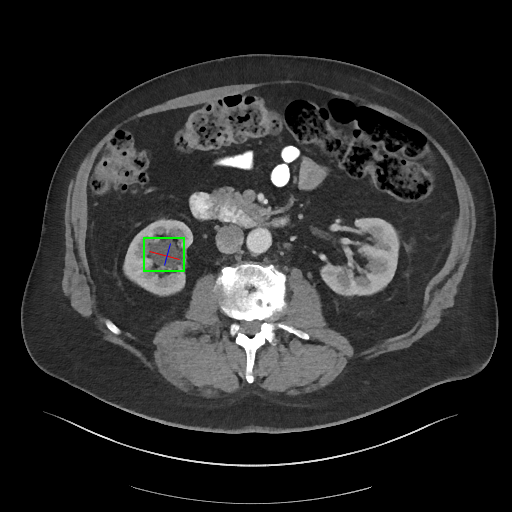}
\end{center} 
\textbf{Ground Truth}: right renal parapelvic cyst. 
\vspace{0.15cm}
\\ \textbf{GPT-4V}: Location: \textcolor{blue}{Right abdomen}, \textcolor{purple}{near the kidney}; Body Part: \textcolor{blue}{Abdomen, kidney}; Type: \textcolor{red}{Mass}; 
Impression: \textcolor{blue}{Well-circumscribed homogenous} \textcolor{red}{mass} \textcolor{blue}{in the right abdomen}, \textcolor{purple}{adjacent to the right kidney.}
\vspace{0.15cm}
\\ \textbf{Gemini Pro Vision}: Location: \textcolor{blue}{Right kidney}; Body Part: \textcolor{blue}{Kidney}; Type: \textcolor{red}{Mass}; Impression: \textcolor{blue}{Right renal} \textcolor{red}{mass}, \textcolor{blue}{likely benign}.
\vspace{0.15cm}
\\ \textbf{LLaVA-Med}: \textcolor{purple}{The image is an axial computed tomography (CT) scan that has been annotated with a bounding box by a radiologist. The bounding box is a rectangular box that encloses the lesion, which is indicated in green. This helps to highlight the area of interest and provides a clear view of the lesion's location, size, and shape}.
\vspace{0.15cm}
\\ \textbf{RadFM}:  \textcolor{red}{\{ Lesion 1 \}}
\end{tcolorbox}
\caption{Comparison of the responses from multi-modal LLMs (using CoT reasoning) for a renal cyst in the right kidney. \textcolor{red}{Red}, \textcolor{blue}{blue} and \textcolor{purple}{purple} fonts denote incorrect, correct, and uncertain descriptions respectively.} 
\label{fig:cot}
\end{figure}



\subsection{Text-Based CoT}

Through Text-Based CoT, the free-text abnormality description generated by a multi-modal LLM should contain the following aspects: Body Part, Location (specific), and Lesion Type. The prompt used for this task was designed to allow the model to concentrate on each aspect individually, thereby optimizing the use of its natural language generation capabilities to produce clinically relevant and informative descriptions. This approach contrasted with the one-shot methods \cite{wei2022chain} that attempted to generate entire reports in a single step without explicit intermediate reasoning. 

``Body Part'' is the larger anatomical region or organ of the body (e.g. liver) where the lesion or abnormality is situated. ``Location'' refers to the specific site within a body part (e.g., Couinaud segment 2 of liver) where the abnormality is located. ``Type'' includes classifications, such as nodule, mass, or enlarged lymph node. The description should be concise and clinically relevant, such that the characteristics of the findings can be pre-filled in the findings section of a radiology report. 

Through experiments, it was observed that LLaVA-Med and RadFM were unable to leverage text-based CoT as shown in Figure \ref{fig:cot}. GPT-4V and Gemini Pro Vision effectively used CoT to provide detailed and relevant descriptions, demonstrating stronger comprehension skills. In contrast, LLaVA-Med, despite being tasked with using CoT, did not produce an analysis related to CoT, focusing instead on the visual elements of the scan, such as the bounding box. RadFM also showed limited capability and offered minimal output, which aligned with findings in literature \cite{kim2023the}. This discrepancy highlighted the architectural or design limitations that hinder certain models from effectively processing input information in a sequential manner. Additional comparative analyses, including those with and without CoT, are detailed in the supplementary material.


\subsection{Fine-Tuning RadFM}

These models have not been specifically fine-tuned for lesion detection on Chest CT scans. Instead, they are often applied in a zero-shot setting, where they are expected to generalize without prior training on the specific task \cite{li2024llava}. To address this,  RadFM was fine-tuned using domain-specific non-overlap data from the DeepLesion subset \cite{yan2019holistic}, to enhance its ability to produce clinically accurate descriptions of CT findings. Fine-Tuning the model required 1 x 80GB A100 GPU, and took approx. 4 days. We believe a major reason for these issues is the lack of public datasets with paired CT studies and detailed descriptions of findings, which are essential for training effective medical imaging models, so we prepossess this dataset fist. The initial dataset, created by radiologists, was not formatted suitably for direct fine-tuning. Therefore, we utilized the GPT-4 API to systematically organize the findings; relevant descriptions of findings and measurements were extracted (sample examples are in the supplementary material). Cases lacking informative descriptions were excluded, resulting in a refined dataset comprising 17,907 descriptions linked with CT images for fine-tuning.
These descriptions served as the ground truth for the fine-tuning process. The enhanced model, designated as RadFM (FT), utilized these datasets. The effectiveness of RadFM (FT) was subsequently assessed using the ``GPTRadScore'' evaluation framework, confirming the enhancements in its performance.

\subsection{GPTRadScore: Evaluation using GPT-4}

``GPTRadScore''  is the cornerstone of our framework, leveraging GPT-4 to replicate the evaluation processes traditionally conducted by radiologists. This system assesses the ability of other multimodal LLMs to generate accurate descriptions for prospectively-identified radiological findings. Specifically, GPT-4 evaluates the accuracy of summaries provided by multimodal LLMs against gold-standard sentences derived from the DeepLesion dataset. Prioritizing criteria most significant to radiologists, the evaluation is segmented into three key aspects: body part, location, and type.
For each category, GPT-4 assigned one of the following categorical scores: ``Correct'', ``Partially Correct'', ``Incorrect'', and ``Not Applicable''. A ``Correct'' score indicated a completely accurate interpretation. ``Partially Correct'' suggested the interpretation captured some aspects accurately, but lacked complete precision or detail. ``Incorrect'' implied the interpretation did not align with the gold-standard at any level. ``Not Applicable'' was used when relevant information was omitted from the description and thus evaluated was not possible. 
Detailed instructions provided to GPT-4, as outlined in the supplementary materials, guided the model to analyze these predictions in a manner akin to clinical judgment. Meanwhile, GPT-4 provided relevant explanations when scoring (see supplementary material) and this capability underscored the advantages of leveraging GPT-4 for complex medical evaluation tasks, where it is crucial to understand detailed anatomical context.

\section{Experiment Setup}
\label{sec_expRes}
\begin{table*}[!t]\small
\centering
\begin{tabular}{ll|cccc|cc}
\toprule
{\textbf{Model}} & {\textbf{}} & 
{\textbf{BLEU\_1}} & {\textbf{BLEU\_2}}&{\textbf{BLEU\_3}}&{\textbf{BLEU\_4}}&\textbf{ROUGE} & \textbf{METEOR} \\
\midrule
\multirow{4}{*}{\textbf{GPT-4V}}      
& bbox CoT          & 0.164 & 0.048 & 0.015 & \bf{0.003} & 0.171 & \bf{0.165} \\
& w/o bbox CoT      & 0.099 & 0.016 & 0.004 & 0.000 & 0.103 & 0.107 \\ 
& bbox w/o CoT      & 0.057 & 0.007 & 0.002 & 0.000 & 0.057 & 0.146 \\ 
& w/o bbox w/o CoT  & 0.022 & 0.002 & 0.000 & 0.000 & 0.029 & 0.071 \\
\midrule
\multirow{4}{*}{\textbf{Gemini Pro Vision}}  
& bbox CoT          & 0.160 & 0.037 & 0.007 & 0.000 & 0.180 & 0.137 \\
& w/o bbox CoT      & 0.116 & 0.025 & 0.006 & 0.000 & 0.137 & 0.108 \\
& bbox w/o CoT      & 0.061 & 0.014 & 0.004 & 0.001 & 0.085 & 0.140 \\
& w/o bbox w/o CoT  & 0.025 & 0.005 & 0.001 & 0.000 & 0.045 & 0.075 \\
\midrule
\multirow{2}{*}{\textbf{LLAVA-Med}}   
& bbox              & 0.024 & 0.002 & 0.000 & 0.000 & 0.025 & 0.062 \\
& w/o bbox          & 0.028 & 0.002 & 0.000 & 0.000 & 0.031 & 0.069 \\
\midrule
\multirow{2}{*}{\textbf{RadFM}}       
& bbox              & 0.096 & 0.011 & 0.000 & 0.000 & 0.075 & 0.095 \\
& w/o bbox          & 0.094 & 0.010 & 0.001 & 0.000 & 0.075 & 0.095 \\
\midrule
\multirow{2}{*}{\textbf{RadFM (FT)}}  
& bbox              & \bf{0.203} & \bf{0.058} & \bf{0.016} & 0.002 & \bf{0.205} & 0.159 \\
& w/o bbox          & 0.187 & 0.052 & 0.015 & 0.001 & 0.195 & 0.151 \\
\bottomrule
\end{tabular}
\caption{Comparative performance of various natural language generation models using BLEU, ROUGE, and METEOR metrics, including fine-tuned RadFM (FT). ``bbox'' meant with bounding box, ``w/o bbox'' meant without bounding box. These metrics, which measured word overlap, showed low scores across the board. This suggested limitations in handling tasks that require deep contextual understanding. This highlighted the need for more sophisticated evaluation methods to gauge true performance.}
\label{tra}
\end{table*}

\subsection{Human Evaluation Process}

To assess the effectiveness of GPT-4 in automatically evaluating findings from multimodal LLMs, we need to establish a human evaluation baseline. This involves comparing AI-generated findings with the ground truth as evaluated by a human expert. We undertook a human evaluation through a structured, collaborative, and iterative process. For this analysis, we randomly selected 100 lesions from a total pool of 500 for each of the five models evaluated: GPT-4V, Gemini Pro Vision, LLaVA-Med, RadFM, and RadFM (FT). Initially, the 500 cases (100 from each model) were analyzed by three graders (PhDs) with  bioinformatics backgrounds. The grading guidelines provided mirror the prompt issued to GPT-4. These assessments were subsequently reviewed and enriched with clinical insights during discussions with a clinician. Any ambiguous findings were collaboratively refined and confirmed, ensuring that the final evaluations were both scientifically robust and clinically relevant. The outcomes of this process is then compared with GPT-4's evaluation of the same report.

\noindent
\subsection{Implementation Details} 
The Advanced 1.5 Pro setting of Gemini Pro Vision was used. LLaVA-Med and RadFM were run using the default configurations. For the model evaluation, we employed the Azure API for GPT-4, configured with a ``temperature'' of 0, ``top\_p'' of 0.95, ``max\_tokens'' of 4000, and the ``model\_version'' set to ``2024-02-15-preview''. 


\subsection{Metrics}  The quality of the generated descriptions were initially assessed using traditional NLG metrics from Huggingface evalaute package, including BLEU, METEOR, and ROUGE. Then, following the approach suggested \cite{zhu2024leveraging}, we conducted an auto-evaluation using GPTRadScore, where the model's predictions were compared against gold-standard annotations. Additionally, these evaluations were compared with assessments conducted by a clinician. The Pearson’s Correlation Coefficient \cite{Pearson1895} between the GPTRadScore and clinician evaluations served as an indicator of GPT-4's reliability for auto-evaluation tasks.

\noindent

\section{Results and Discussion}
\subsection{Traditional NLG Metrics Analysis}
\noindent\textbf{Results:}
Table \ref{tra} evaluates multi-modal LLMs using traditional NLG metrics, and differentiates their performance in scenarios with and without bounding box constraints. RadFM (FT) bbox model exhibited outstanding performance across all NLG metrics, and particularly excelled at structural alignment and linguistic matching. GPT-4V bbox with CoT and Gemini Pro Version bbox with CoT also performed well; GPT-4V bbox with CoT achieved the highest METEOR score of 0.165. Conversely, the non-fine-tuned versions of RadFM, LLAVA-Med, and GPT-4V without CoT exhibited substantially lower performance. This decline is likely due to  significant domain shift of the test dataset (radiology reports) in contrast to the model training dataset. Despite this discrepancy, the experiment was setup to expose the limitations of traditional NLG metrics.  

\noindent\textbf{Limitations of traditional metrics:}
While traditional NLG metrics are valuable for assessing linguistic quality, they do not fully capture the clinical relevance of the generated descriptions. In clinical settings, the priority lies in the factual accuracy and clinical relevance of generated summary over mere linguistic fidelity. This highlights the need for more robust evaluation methods that better mirror the utility in medical contexts.



\subsection{Correlation between Clinician, GPTRadScore, Traditional Metrics}

\begin{figure*}[!t]
    \centering
    \includegraphics[trim={0cm 0 4cm 5cm},clip,width=0.3\textwidth]{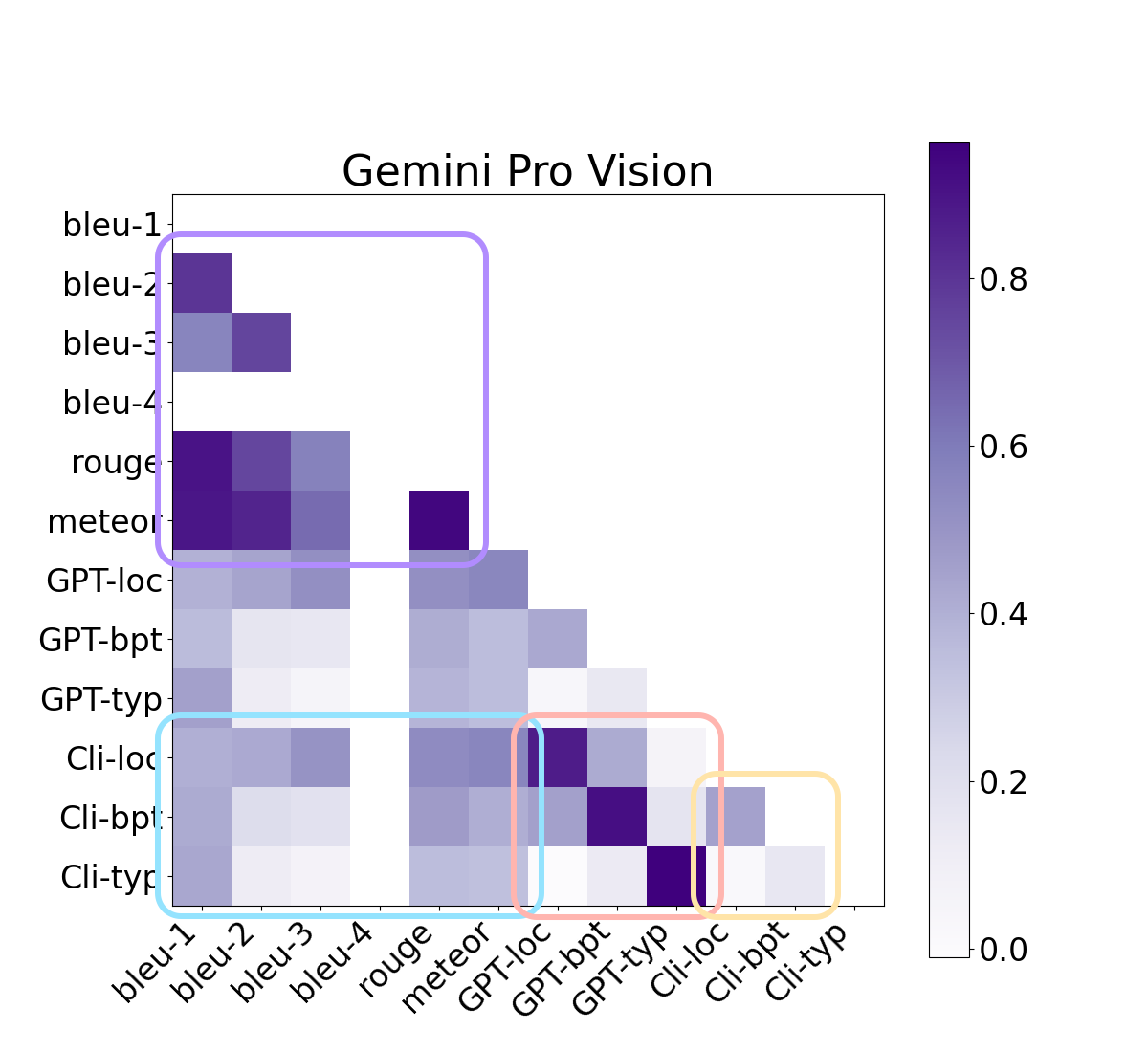}
    \includegraphics[trim={0cm 0 4cm 5cm},clip,width=0.3\textwidth]{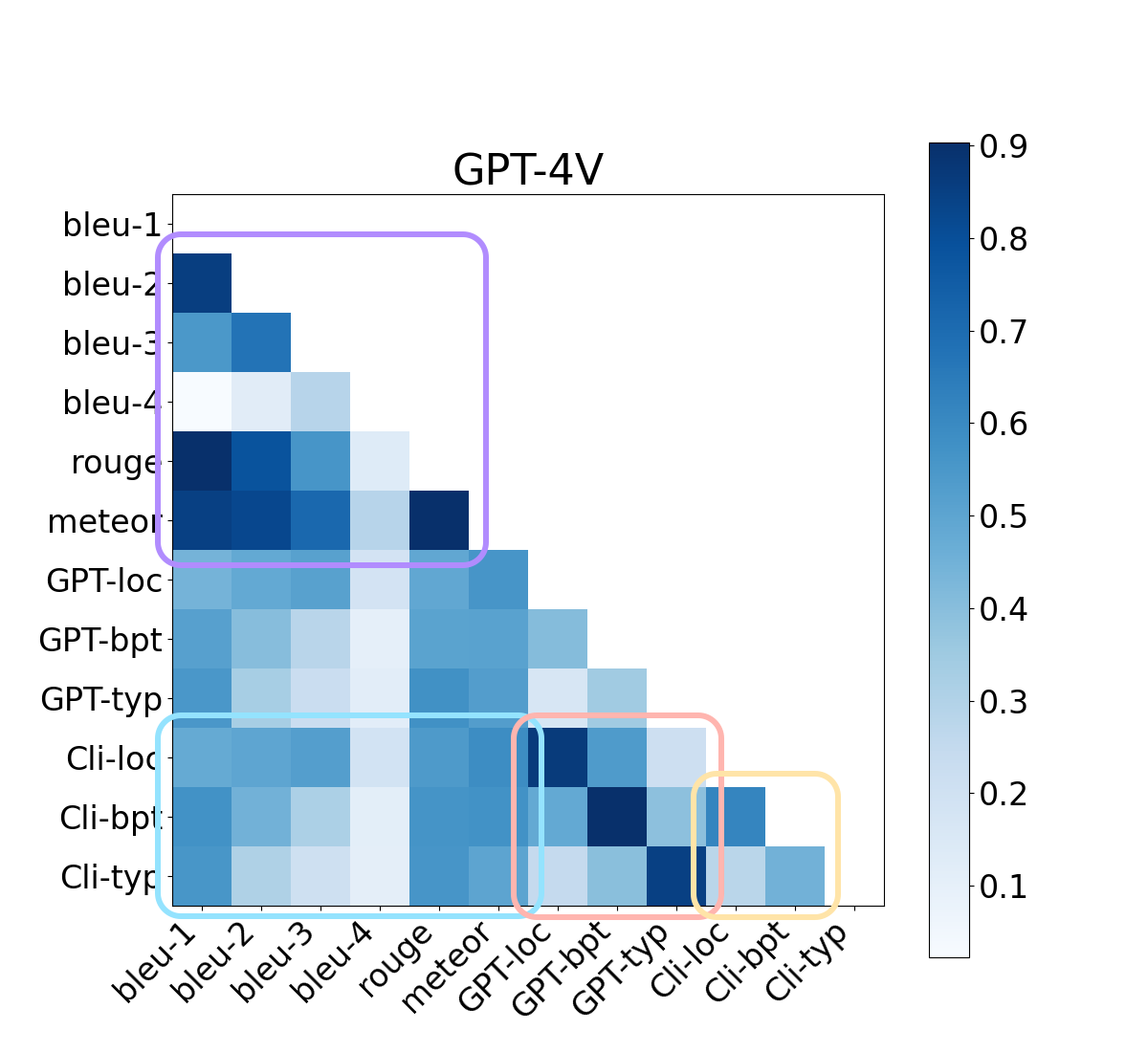}
    \includegraphics[trim={0cm 0 4cm 5cm},clip,width=0.3\textwidth]{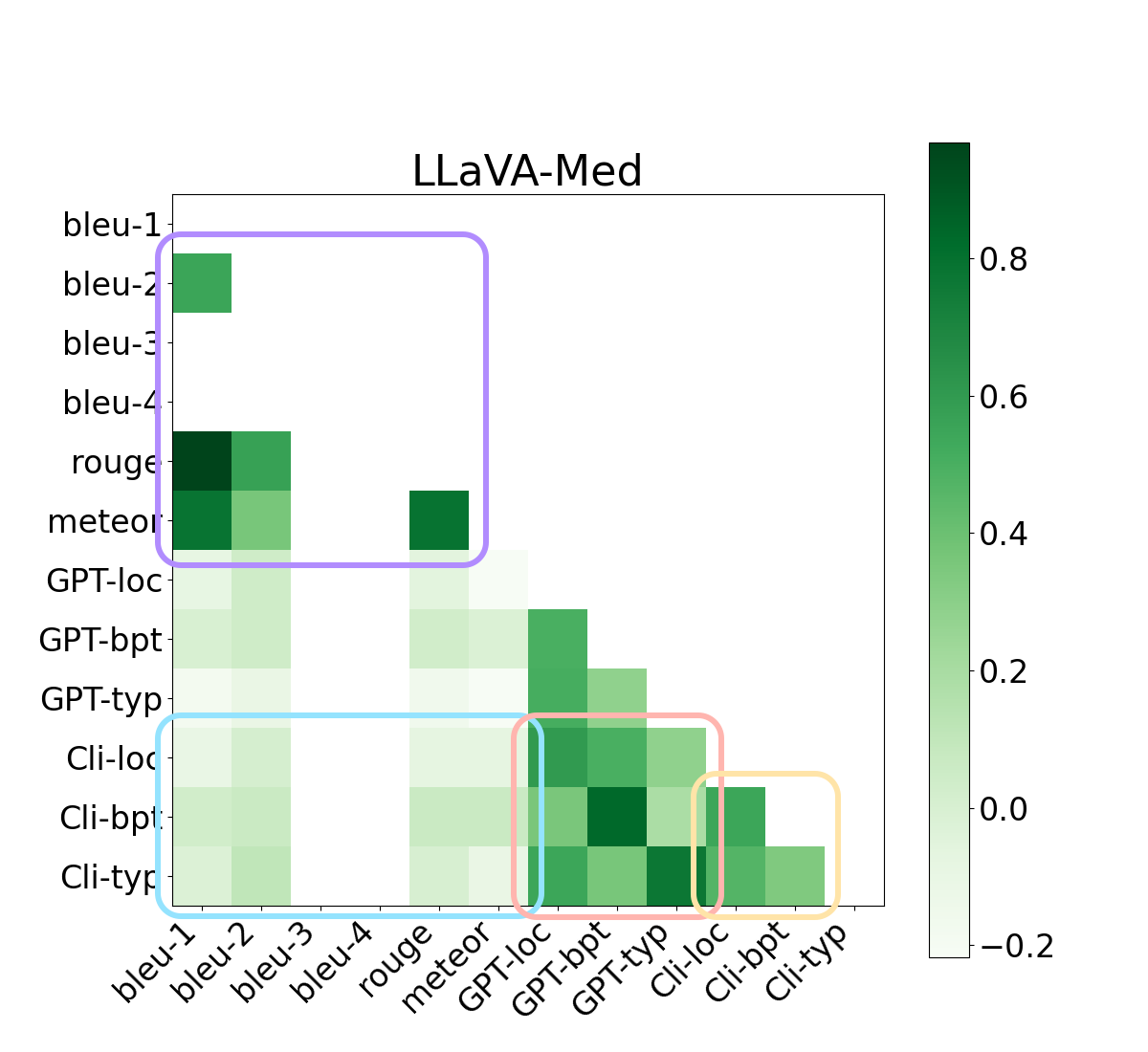}
    \includegraphics[trim={0cm 0 4cm 5cm},clip,width=0.3\textwidth]{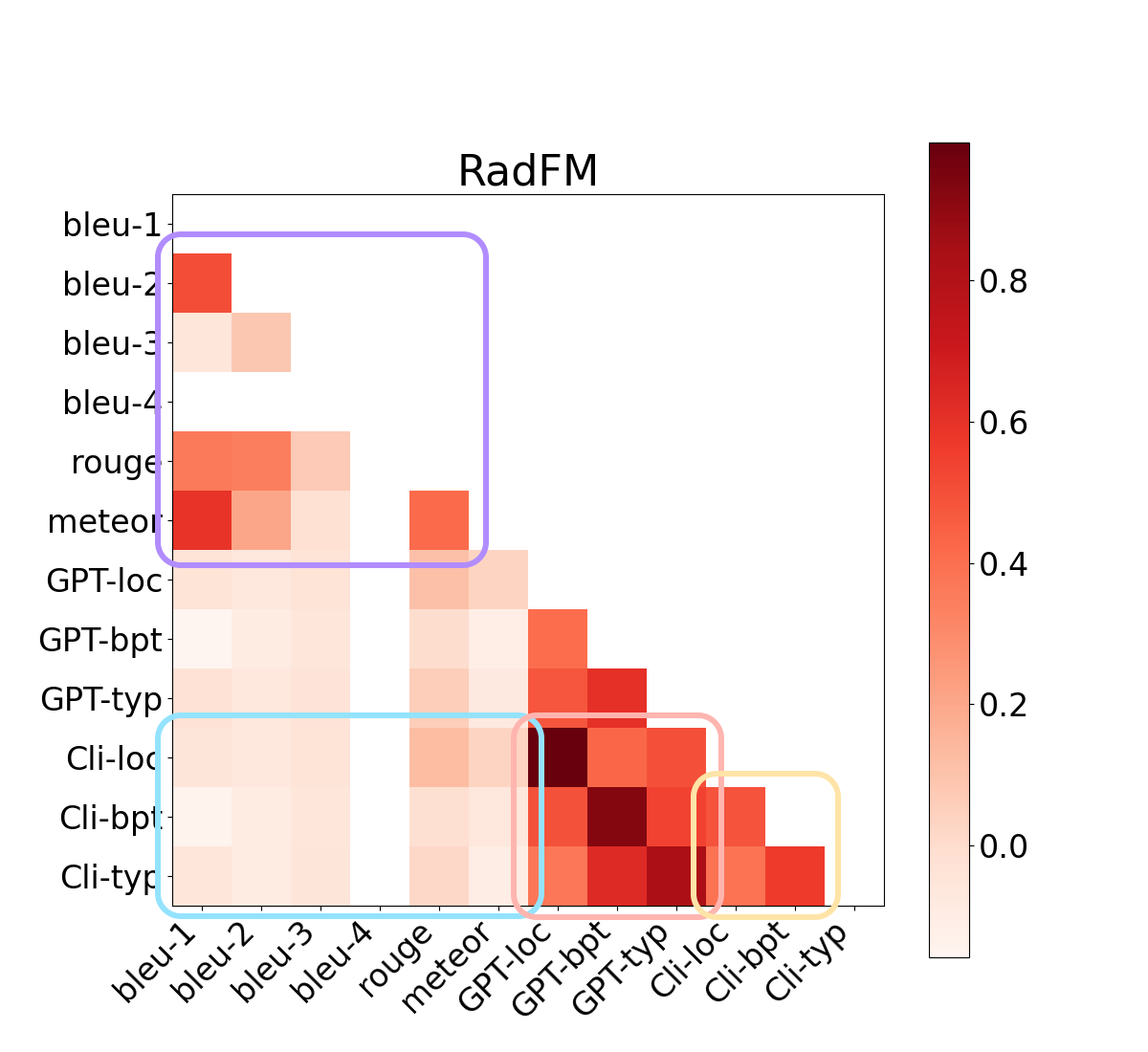}
    \includegraphics[trim={0cm 0 4cm 5cm},clip,width=0.3\textwidth]{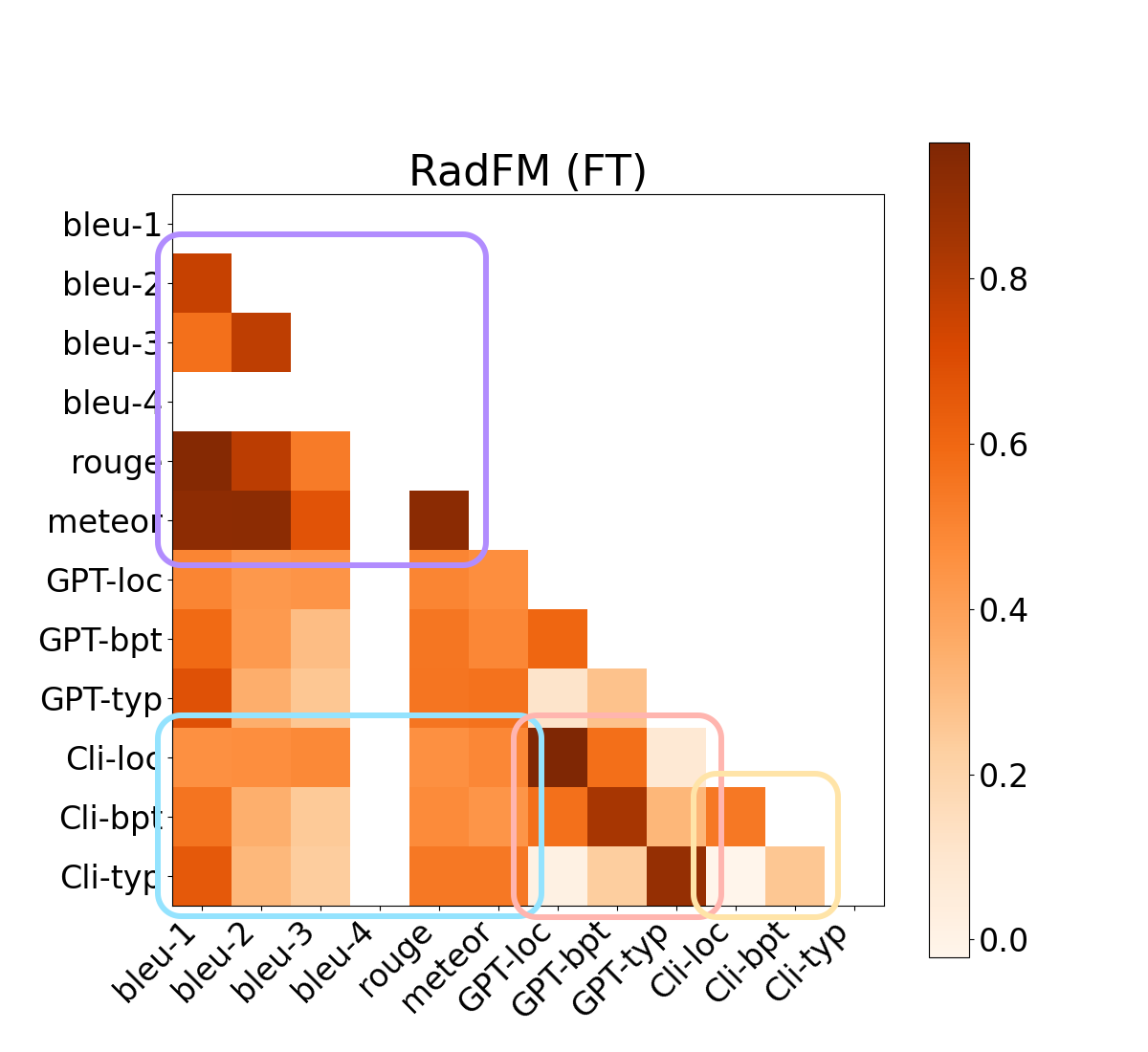}
    \caption{Heatmap of pairwise Pearson’s Correlation Coefficient among various grading scores; traditional metrics, Clinician evaluations and GPTRadScore for Gemini Pro Vision, GPT-4V, LLaVA-Med, RadFM, and RadFM (FT). Color intensity indicates the strength of correlation, with darker shades representing higher correlation.}
    \label{fig:corr_plot}
\end{figure*}

Figure \ref{fig:corr_plot} shows the correlation between the various metrics computed based on the ground truth evaluation, GPTRadScore, and traditional NLG metrics.
Traditional NLG metrics like BLEU and METEOR demonstrate strong correlations amongst themselves (purple box), particularly at lower levels of precision like BLEU-1 and BLEU-2. This indicated a consistency in evaluating the linguistic quality of generated texts at these levels. However, at higher precision levels (BLEU-3 and BLEU-4), these correlations significantly weaken, particularly for LLaVA-Med, where scores frequently register at zero and indicate no correlation. This pattern again reflects the limitations of traditional metrics in evaluating complex sentence structures typical in radiology reports.


Furthermore, the decomposition of the description into specific aspects (location, body part, and lesion type) revealed insightful patterns (peach box). These aspects showed a lack of strong correlation with one another, and other pairings also displayed no significant correlation. These observations affirmed the efficacy of the approach in dissecting the findings into their granular elements, such that the distinct parts of report quality can be assessed independently. 

Comparing traditional metrics with the ground truth evaluation showed a weak correlation, suggesting that these metrics may not serve as reliable indicators of clinical accuracy for radiology reports (blue box). This highlighted a potential gap in utilizing NLG metrics for assessing the clinical relevance of generated reports, pointing to the necessity for domain-specific evaluation methods. 


Lastly, the comparison between GPTRadScore and the ground truth evaluation showed the strength of our framework (pink box), and summarized in Table \ref{tab:corr_scores}. The results showed a strong correlation with ground truth, suggesting that GPTRadScore  closely aligned with the clinical assessment paradigms utilized by radiologists. This observation underscored the potential of LLMs like GPT-4 in accurately mirroring radiologists' evaluations, offering promise for automating assessment with a high degree of fidelity to clinical standards.

\begin{table}[!h]
\centering
\resizebox{0.48\textwidth}{!}{
\begin{tabular}{@{}l|ccc|cc@{}}
\toprule
                    & Location & Body Part & Type  & Avg.            & p-value   ~ \\ 
\midrule
~GPT-4V             & 0.86     & 0.90      & 0.84  & 0.87 $\pm$ 0.02 & $<$0.001  ~ \\
~Gemini Pro Vision  & 0.87     & 0.91      & 0.96  & 0.91 $\pm$ 0.03 & $<$0.001  ~ \\
~LLaVA-Med          & 0.59     & 0.83      & 0.76  & 0.75 $\pm$ 0.10 & $<$0.001  ~ \\
~RadFM              & 0.99     & 0.92      & 0.82  & 0.90 $\pm$ 0.07 & $<$0.001  ~ \\
~RadFM (FT)         & 0.96     & 0.83      & 0.89  & 0.89 $\pm$ 0.05 & $<$0.001  ~ \\
\bottomrule
\end{tabular}
}
\medskip
\caption{Correlation scores between the clinician and GPT-4 grading of reports.}
\label{tab:corr_scores}
\end{table}





\subsection{GPTRadScore Evaluation}

\begin{figure*}[!ht]
\centering
\includegraphics[width=0.79\textwidth]{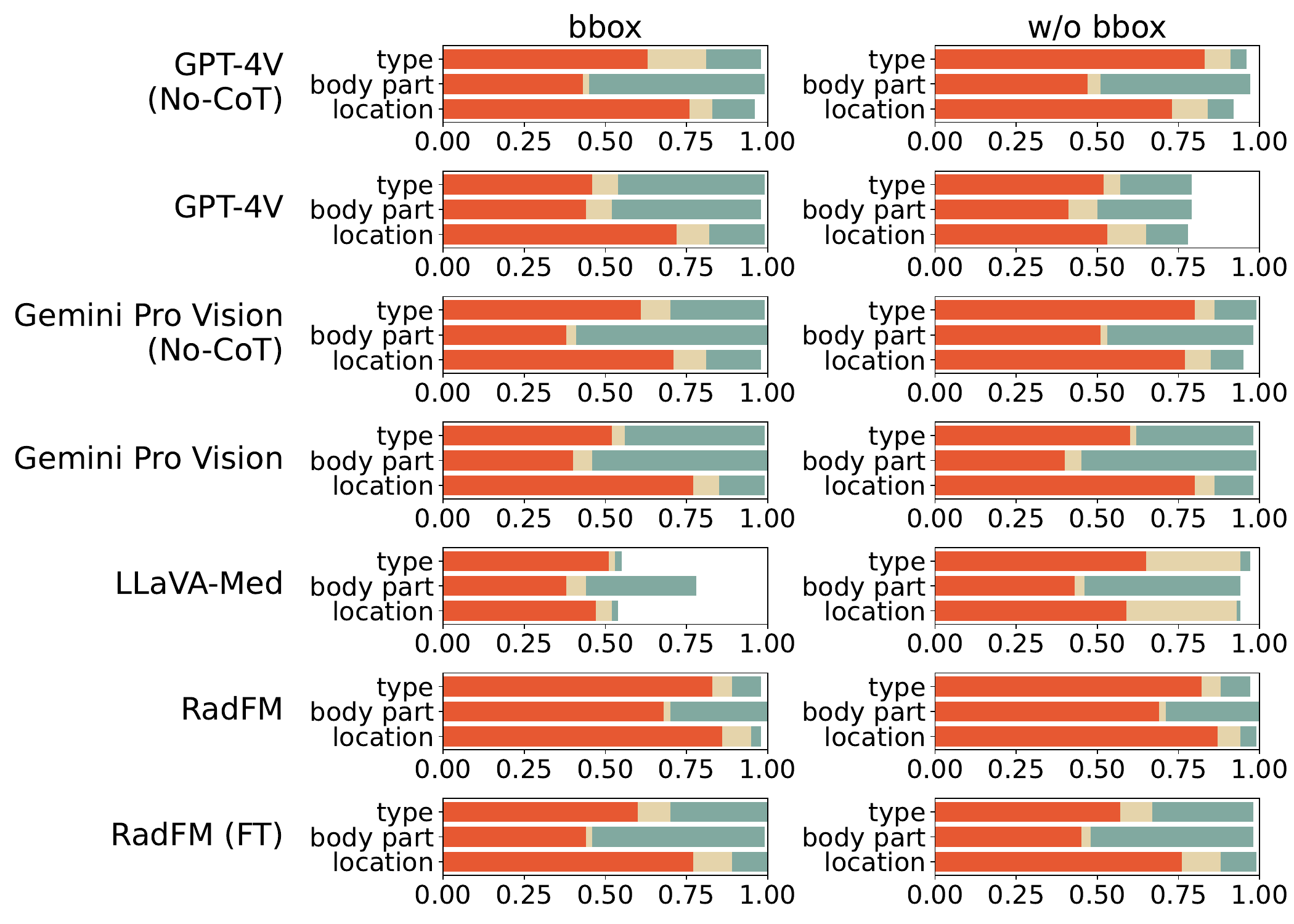}
\caption{
Comparison of results of abnormality characterization by GPT-4V, Gemini Pro Vision, LLaVA-Med, RadFM, and RadFM (FT) with bounding boxes (bbox) vs. without bounding boxes (w/o bbox). Color mapping = \{orange: `Incorrect', beige: `Partially Correct', teal: `Correct', white: `Not Applicable'\}. $x$-axis denotes scores $\{\ x\in \mathbb{R} \ | \ 0 < x < 1\ \}$, $N$ = 500 samples.
}
\label{fig:human_eval}
\end{figure*}

Due to the strong correlation between the GPTRadScore and clinician evaluations, GPTRadScore was employed to assess the predicted findings against the ground truth for all 500 lesions for three categories: location, body part, and type of abnormality. Figure \ref{fig:human_eval} displays the grading scores across four configurations: with and without lesion bounding boxes in the CT slice, and with and without text-based CoT. Notably, LLaVA-Med and RadFM do not support text-based CoT processes; thus, this figure exclusively presents the use of CoT in the GPT-4V and Gemini Pro Vision models. \textit{Bounding boxes consistently enhanced identification of body part and location across all models, thereby indicating a dependency on strong visual cues for accurate recognition}. For example, GPT-4V and Gemini Pro Vision performed better with bounding boxes, particularly for accurate body part identification. 

To better illustrate the impact of CoT reasoning, besides this figure, we have included correct classification percentages for GPT-4V and Gemini Pro Version with bounding boxes in Table \ref{tab:classification_percentages_boxes_only}. \textit{CoT significantly boosts type classification accuracy}, with improvements of 28.1\% for GPT-4V (from 16.5\% to 44.6\%) and 14.39\% for Gemini Pro Vision (from 28.78\% to 43.17\%). However, its influence on location and body part classifications is less pronounced and may even hinder performance in some cases. This suggests that the structured reasoning provided by CoT particularly benefits complex decision-making tasks requiring nuanced interpretation and detailed contextual understanding. Conversely, for simpler tasks like identifying locations or body parts, a direct approach without CoT tends to be more effective. Overall, the enhanced performance of  CoT indicates that its architecture is well-suited to sequential reasoning, akin to a radiologist’s thought process, thus leading to more accurate descriptions of CT findings.

\begin{table}[!ht]\small
\centering

\begin{tabular}{l|cc|cc}
\toprule
\multirow{2}{*}{\textbf{Category}} & \multicolumn{2}{c|}{\textbf{GPT-4V}} & \multicolumn{2}{c}{\textbf{Gemini Pro Vision}} \\

                                   & \textbf{CoT} & \textbf{w/o CoT} & \textbf{CoT} & \textbf{w/o CoT} \\
\hline
Location                           & 17.1\%           & 13.0\%             & 14.02\%          & 17.27\%            \\

Body Part                          & 46.4\%           & 53.5\%             & 53.51\%          & 58.64\%            \\

Type                               & 44.6\%           & 16.5\%             & 43.17\%          & 28.78\%            \\
\bottomrule
\end{tabular}
\caption{Correct Classification Percentages for GPT-4V and Gemini Pro Version with Bounding Boxes.}
\label{tab:classification_percentages_boxes_only}
\end{table}

\textit{Among the non-fine-tuned models, GPT-4V and Gemini Pro Vision excel in medical imaging tasks, which is likely attributed to their extensive pre-training on diverse datasets.} Specifically, in tasks requiring identification within  bounding boxs, GPT-4V scores 46.4\% in type recognition and 53.5\% in body part identification. Gemini Pro Vision follows closely with scores of 44.6\% and 43.17\%, respectively. Despite outperforming other models, both GPT-4V and Gemini Pro Vision have room for improvement in lesion location accuracy, with scores of 17.1\% and 14\% respectively.
In contrast, models like LLaVA-Med and RadFM demonstrate significantly weaker performance, particularly in tasks without spatial cues. For example, The LLaVA-Med bbox model achieves accuracies of 1.81\% for identifying lesion locations and types, and 33.73\% for recognizing body parts. These models struggle to generalize from their training data, highlighting significant challenges in adapting AI to real-world medical tasks. The suboptimal performance of models stems from the fact that these models have not been specifically fine-tuned for lesion detection on Chest CT scans \cite{li2024llava}.

To prove this, we fine-tuned the RadFM model by using domain-specific, non-overlapping data. RadFM (FT) exhibited improvement across all three categories compared to the standard RadFM. Especially when bounding boxes were employed, the location accuracy rose from 3.41\% to 12.8\%, body part accuracy increased from 29.12\% to 53\%, and type accuracy improved from 9.24\% to 30\%. \textit{This indicated that fine-tuning and targeted optimizations with  data effectively address specific weaknesses in model performance, suggesting a pathway for further enhancing the reliability of multi-modal LLMs for medical imaging.}

\section{Conclusion}

In summary, we proposed the novel ``GPTRadScore'' framework for automatically evaluating AI-generated descriptions of findings prospectively identified in CT exams. These descriptions  were intended to be pre-filled into the radiology reports' findings section. Four multi-modal LLMs were tested for the ability to generate a description of a CT-based finding when fed with an input CT slice. GPT-4V and Gemini Pro Vision notably outperformed other recent multi modal LLMs in accurately predicting lesion characteristics. Bounding boxes outlining the lesions in the CT slices provided strong visual cues and consistently helped these multi-modal LLMs to identify the body part and location correctly. GPTRadScore auto-evaluation results demonstrated a strong correlation with clinician assessments as measured by Pearson's correlation coefficient. Our evaluation highlighted specific weaknesses in various multi-modal LLMs, primarily due to the dataset limitations that these models were trained on. By fine-tuning RadFM on domain-specific data, significant enhancements substantially improve the utility of multi-modal LLMs in radiology.

\section{Limitions}

One limitation of our study is the lack of investigation into prompt engineering. We utilized the prompts recommended by the model developers, assuming these would optimize performance. However, more meticulously crafted prompts could potentially yield better outcomes. This reliance on predefined prompts mirrors the early days of image-based pattern recognition, suggesting that just as image recognition evolved to require less manual intervention, prompt engineering may also become more automated and effective in the future. To address this, future research could explore automated prompt generation techniques or machine learning algorithms that optimize prompt selection based on task specifics and data context.

Another limitation involves the cost and practicality of implementing such advanced AI models in clinical settings. As LLMs continue to evolve, the associated deployment costs are expected to decrease, making the technology more accessible and feasible for wider implementation. Our model shows a key direction that can be used in the future to further this progress. To mitigate high costs and enhance practicality, solutions such as developing computationally efficient models, utilizing cloud-based deployments, forming partnerships with technology providers, and initiating pilot projects could be pursued. These strategies can demonstrate the benefits of AI technologies and support broader adoption, aligning with the ongoing advancements and cost reductions in the field of LLMs.

\section{Ethical Statement}

The data used in this study are fully anonymized, and comply with the Health Insurance Portability and Accountability Act (HIPAA). Use of the data was approved by the Institutional Review Board (IRB), and the requirement for informed consent was waived. For the generation task, publicly available images from DeepLesion were used. For the evaluation task, we accessed GPT-4, an online large language model, via Microsoft Azure services to guarantee secure and privacy-compliant data handling.


\bibliography{references}
\clearpage
\onecolumn
\section{Appendix}
\label{sec:appendix}

\begin{table*}[!h]\small
\centering

\begin{tabular}{p{0.15\linewidth}p{0.8\linewidth}}
\hline
\textbf{Scenario} & \textbf{Prompt} \\ \hline
w/o CoT, bbox & This image is with a bounding box created by a radiologist. Imagine you are a radiologist. Generate a short radiological impression based on this image. \\ \hline
w/o CoT, w/o bbox & Imagine you are a radiologist. Generate a short radiological impression based on this image. \\ \hline
CoT, bbox & Please describe this image in detail, which is with a bounding box created by a radiologist. When describing this image, please point this: 1. Location: Refers to the specific area where the lesion is found. For example: the outer edge of the lower left lung; 2. Body Part: Indicates the larger region of the body where the lesion is located. For example: lung; 3. Types, which include general terms (e.g., nodule, mass) and more specific ones (e.g., liver mass); 4. Impression: Summarize the most significant findings. \\ \hline
CoT, w/o bbox & Please describe this image in detail. When describing this image, if this image contains a lesion, please point this: 1. Location: Refers to the specific area where the lesion is found. For example: the outer edge of the lower left lung; 2. Body Part: Indicates the larger region of the body where the lesion is located. For example: lung; 3. Types, which include general terms (e.g., nodule, mass) and more specific ones (e.g., liver mass); 4. Impression: Summarize the most significant findings. \\ \hline
Evaluation & This task involved evaluating the accuracy of a predicted diagnostic interpretation (pred\_result) against a ground truth description (gt\_text) and its expanded terms in the context of medical imaging findings. What's Checked:
Location: If the model finds the problem in the right spot.
Body Part: If the model correctly names which part of the lesion located.
Type: If the model accurately describes what kind of issue it sees (like a nodule or mass).
Grading Terms:
Correct: The model's guess is just right.
Partially Correct: The model's guess is somewhat right but lacks full accuracy or completeness.
Incorrect: The model's guess doesn't match the correct answer at all.
Not Applicable: The model's guess omits relevant information and thus cannot be evaluated
\\ \hline
\end{tabular}
\caption{Prompts for CT Reporting Generation and Evaluation}
\end{table*}

\begin{table*}[!ht]\small
\centering
\begin{tabular}{p{8.5cm} p{3.5cm} p{2cm}}
\midrule
\textbf{Original Sentence} & \textbf{Current Description} & \textbf{Current Size} \\
\toprule
There is no mediastinal adenopathy however there is a nodule in the prevascular space measuring BOOKMARK ( 1.8 cm x 1.0 cm ) ( series 3 , image 88 ) .
 & There is no mediastinal adenopathy; however, there is a nodule in the prevascular space.
& 1.8 cm x 1.0 cm \\
\midrule
Smaller retroperitoneal nodules and masses for example iliac artery OTHER\_BMK ( 1.6 cm x 1.4 cm ) , prior exam was OTHER\_BMK ( 3.4 cm x 1.8 cm ) and left internal iliac BOOKMARK ( 2.0 cm x 1.2 cm ) , prior exam OTHER\_BMK was ( 5.1 cm x 4.4 cm )
 & Retroperitoneal nodules and masses, for example left internal iliac. & 2.0 cm x 1.2 cm \\

\bottomrule
\end{tabular}
\caption{This table illustrates the methodical extraction and analysis of radiological data from medical imaging reports, specifically focusing on measurements based on the RECIST criteria, referred to as 'bookmarks.' It details how the Current Description and Current Size are extracted from the original text using the GPT-4 API.}
\label{tab:medical_imaging}
\end{table*}

\begin{table*}[h!]\small
    \centering
    
    \begin{tabular}{llp{11cm}}
        \hline
        \multicolumn{3}{p{15cm}}{\textbf{gt\_text}: Right posterior hemithorax mass} \\
        \hline
        \multicolumn{3}{p{15cm}}{\textbf{pred\_result}: The image is a computed tomography (CT) scan of the chest, taken during the arterial phase. It shows a mass in the right upper lobe of the lung, which appears to have lobulated contours and blurry margins. The mass has a central caliber and a diameter of around 5 cm. Additionally, there is a solid lesion measuring approximately 3 x 1.5 cm at the posterior wall of the trachea.} \\
        \hline
        \textbf{Criterion} & \textbf{Evaluation} & \textbf{Explanation} \\
        \hline
        Location & Partially Correct & While the prediction does identify a 'right' sided issue and a finding in the 'posterior' aspect, it describes the mass being in the 'right upper lobe', whereas the ground truth indicates a 'right posterior hemithorax' location. \\
        \hline
        Body Part & Correct & The prediction correctly identifies the 'lung' and 'chest' as the body parts involved, which is consistent with the ground truth. \\
        \hline
        Type & Correct & The prediction correctly identifies a 'mass', which is consistent with the ground truth. \\
        \hline
        Attributes & Partially Correct & The prediction describes some attributes of the mass (e.g., 'lobulated contours', 'blurry margins', 'central caliber', 'diameter of around 5 cm') but these are not specified in the ground truth. The prediction is however incorrect in describing the lesion at the 'posterior wall of the trachea' because this is not mentioned in the ground truth. \\
        \hline
       
    \end{tabular}
    \caption{Evaluation from GPT-4 for LLaVA-Med Generated Findings. The \textit{gt\_text} represents the ground truth, while \textit{pred\_result} contains the sentences generated by LLaVA-Med. }
    \label{tab:llava_evaluation}
\end{table*}

\begin{table*}[h!]\small
    \centering
    
    \begin{tabular}{llp{11cm}}
        \hline
        \multicolumn{3}{p{15cm}}{\textbf{gt\_text}: destructive left iliac bone ilium mass with extraosseous mass invading the adjacent iliopsoas and gluteus minimus muscles} \\
        \hline
        \multicolumn{3}{p{15cm}}{\textbf{pred\_result}: The image is a pre-operative computed tomography (CT) scan of the pelvis. It shows multiple osteolytic sacral and pelvic lesions, as well as a large anterior sacral wall defect and soft tissue invasion. The sacral region appears to have irregular bone destruction, and there are several smaller lesions scattered throughout the pelvis. The largest lesion is located in the left sacral bone, with a diameter of 27 mm.} \\
        \hline
        \textbf{Criterion} & \textbf{Evaluation} & \textbf{Explanation} \\
        \hline
        Location & Partially Correct & The GT mentions 'left iliac bone ilium' while the prediction mentions the 'sacral and pelvic' region. The iliac bone is a part of the pelvic bone, so this is partially correct. \\
        \hline
        Body Part & Correct & The ground truth mentions 'iliac bone ilium' and 'iliopsoas and gluteus minimus muscles.' The prediction mentioned 'sacral and pelvic' bone which includes the body parts in GT, so it is correct. \\
        \hline
        Type & Incorrect & The prediction indicates that there are multiple osteolytic lesions while the GT mentions a destructive mass. There is a mismatch here. An osteolytic lesion is different from a destructive mass. \\
        \hline
        Attributes & Incorrect & While the prediction provides a measure of the 'largest lesion' and mentions 'several smaller lesions,' the ground truth does not mention size or number of masses but describes an 'extraosseous mass invading adjacent muscles,' which the prediction does not detail. \\
      \hline
    \end{tabular}
    \caption{Evaluation from GPT-4 for RadFM Generated Findings. The \textit{gt\_text} represents the ground truth, while \textit{pred\_result} contains the sentences generated by RadFM.}
    \label{tab:radfm_evaluation}
\end{table*}

\begin{table*}[h!]\small
    \centering
    
    \begin{tabular}{llp{11cm}}
        \hline
        \multicolumn{3}{p{15cm}}{\textbf{gt\_text}: destructive left iliac bone ilium mass with extraosseous mass invading the adjacent iliopsoas and gluteus minimus muscles} \\
        \hline
        \multicolumn{3}{p{15cm}}{\textbf{pred\_result}: The image is a pre-operative computed tomography (CT) scan of the pelvis. It shows multiple osteolytic sacral and pelvic lesions, as well as a large anterior sacral wall defect and soft tissue invasion. The sacral region appears to have irregular bone destruction, and there are several smaller lesions scattered throughout the pelvis. The largest lesion is located in the left sacral bone, with a diameter of 27 mm.} \\
        \hline
        \textbf{Criterion} & \textbf{Evaluation} & \textbf{Explanation} \\
        \hline
        Location & Partially Correct & The GT mentions 'left iliac bone ilium' while the prediction mentions the 'sacral and pelvic' region. The iliac bone is a part of the pelvic bone, so this is partially correct. \\
        \hline
        Body Part & Correct & The ground truth mentions 'iliac bone ilium' and 'iliopsoas and gluteus minimus muscles.' The prediction mentioned 'sacral and pelvic' bone which includes the body parts in GT, so it is correct. \\
        \hline
        Type & Incorrect & The prediction indicates that there are multiple osteolytic lesions while the GT mentions a destructive mass. There is a mismatch here. An osteolytic lesion is different from a destructive mass. \\
        \hline
        Attributes & Incorrect & While the prediction provides a measure of the 'largest lesion' and mentions 'several smaller lesions,' the ground truth does not mention size or number of masses but describes an 'extraosseous mass invading adjacent muscles,' which the prediction does not detail. \\
      \hline
    \end{tabular}
    \caption{Evaluation from GPT-4 for RadFM Generated Findings. The \textit{gt\_text} represents the ground truth, while \textit{pred\_result} contains the sentences generated by RadFM.}
    \label{tab:radfm_evaluation}
\end{table*}
\begin{table}[!ht]\small
\begin{tabular}{lp{2cm}p{4cm}p{6cm}}
\toprule
Model &gt\_text  & CoT & w/o CoT \\ 
\midrule       
Gemini & \multirow{2}{2cm}{right renal hypodensity; cyst}   & Location: Right kidney; Body Part: Kidney; Type: Cyst; Impression: Right renal cyst.                                                                                                                                                                                                                                                                                                                                                                                                                                                            & There is a 5 mm non-obstructing calculus in the mid ureter on the right.                                                                                                                                                                                                                                                                                                                     \\ \cmidrule(r){1-1} \cmidrule(l){3-4}
GPT-4V &                                                    & Location: Right lower quadrant of the abdomen, Body Part: Appendix, Type: Appendicolith, Impression: 'Hyperdense focus within the appendix suggestive of appendicolith                                                                                                                                                                                                                                                                                                                                         & There is a well-circumscribed, low-density lesion in the left kidney, which is highlighted by the bounding box. This finding is suggestive of a simple renal cyst. No features suggestive of malignancy are noted in the visualized portion of the abdomen. Further evaluation with ultrasound can be done for confirmation and to exclude any complex features. \\ \midrule
Gemini & \multirow{2}{2cm}{large necrotic right hepatic mass} & Location: Right lower quadrant of the abdomen; Body Part: Abdomen; Type: Mass; Impression: Right lower quadrant abdominal mass, likely representing an abscess.                                                                                                                                                                                                                                                                                                                                                                                                                   & There is a 3.8 x 3.3 x 4.1 cm enhancing mass in the mid right kidney. The mass demonstrates central low attenuation. There is no evidence of perinephric extension or lymphadenopathy."                                                                                                                                                                                                      \\ \cmidrule(r){1-1} \cmidrule(l){3-4}
GPT-4V &                                                    &Location: Right upper quadrant of the abdomen, Body Part: Liver, Type: Lesion,  Impression: Well-circumscribed hypodense lesion in the right lobe of the liver. & The axial CT image of the abdomen with contrast demonstrates a well-circumscribed lesion in the left kidney, as demarcated by the bounding box. The lesion shows homogeneous density and does not appear to involve the renal hilum or adjacent structures. There are no signs of local invasion or distant metastasis. Further evaluation with dedicated renal protocol imaging and potential urology consultation is recommended for characterization and management of the renal lesion. No other significant abnormalities are noted in the visualized portions of the abdomen.    \\
\bottomrule
\end{tabular}
\caption{Generated descriptions with and without the CoT by Gemini and GPt-4V models.}
\end{table}

\end{document}